Original Paper

# Machine-learning-based investigation on the inclusion of location and weather parameters, dataset recalibration, feature selection-based training, and model performances in classifying binary and multiclass behavior outcomes of children with PIMD/SMID

Von Ralph Dane Marquez Herbuela[1], Tomonori Karita[1*], Yoshiya Furukawa[2], Yoshinori Wada[1], Yoshihiro Yagi[1], Shuichiro Senba[3], Eiko Onishi[3], Tatsuo Saeki[3]

[1]Department of Special Needs Education, Graduate School of Education, Ehime University, Japan
[2]Graduate School of Humanities and Social Sciences, Hiroshima University, Japan
[3] DigitalPia Co., Ltd., Japan

*Corresponding Author:
Tomonori Karita, PhD
Department of Special Needs Education, Graduate School of Education, Ehime University, Bunkyo-cho 3, Matsuyama, 790-8577, Japan
Phone: +81-(0)89-927-9517
Email: karita.tomonori.mh@ehime-u.ac.jp

## Abstract

**Background**

Recently, the importance of weather parameters and location information to better understand the context of the communication of children with profound intellectual and multiple disabilities (PIMD) or severe motor and intellectual disorders (SMID) has been proposed. However, an investigation on whether these data can be used to classify their behavior for system optimization aimed at classifying their behavior for independent communication and mobility has not been done. Thus, this study investigates whether recalibrating the datasets including either minor or major behavior categories or both, combining location and weather data and feature selection method training (Boruta) would allow more accurate classification of behavior discriminated to binary and multiclass classification outcomes using eXtreme Gradient Boosting (XGB), support vector machine (SVM), random forest (RF), and neural network (NN) classifiers. Multiple single-subject face-to-face and video-recorded sessions were conducted among 20 purposively sampled 8 to 10 -year old children diagnosed with PIMD/SMID or severe or profound intellectual disabilities and their caregivers. Datasets with a minor, major, and both behavior categories combined with environmental data in classes 2 and 3 had the highest accuracy rates among all the dataset combinations (above 70%, P<.001). The use of NN in binary (69%), SVM (69%), and RF (69%) in class 3 and Boruta-trained dataset with environment data in binary (72%) and non-Boruta dataset with environment data in class 3 (72%) also had significantly higher mean classification accuracy rates. Our study demonstrated the feasibility of classifying the behavior of children with PIMD/SMID and relatively higher accuracy rates can be obtained when environment data were combined with any behavior dataset combination. Moreover, the improvement in the overall classification accuracy rates is also dependent on the interaction between the classifier and classes and the interactions among dataset, feature selection, and classes.

# 1. Introduction
## 1.1. Children's behavior and weather variables

There has been substantial evidence on the relationship between weather parameters and behaviors and affective states of typically developing children across settings and seasons. Temperature, humidity, and barometric or atmospheric pressure were reportedly associated with levels of mood dimensions, feelings, and scores of concentration, performance, and activities of typically developing children [1]. Lagace-Seguin and Coplan(2001) with d'Entremont (2005) reported that increasing humidity is negatively associated with prosocial behavior and positively correlated with irritability [2,3]. Solar radiation has also been found to have a positive association with the level of positive feeling [2]. A unanimous result of a more recent study by Ciucci et al. (2011) also revealed that the levels of frustration, sadness, and aggressive behavior among children increase as humidity increases, and decrease as solar radiation increases [4]. In a follow-up study, Ciucci et al. (2013) concluded that humidity, in particular, whether indoor or outdoor, was found to be an important weather variable that influences children's behavior and affective states in any season [1]. Further, more recently, Harrison et al. (2017) and Kharlova et al. (2020) have also reported that low levels of physical activity among children were found to be associated with increased precipitation and wind speed and decreased visibility and fewer hours of daylight [5,6]. Despite the considerable number of studies among typically developing children, studies involving children with neurological functioning impairments that affect communication or those who have physical and/or motor disabilities are unexpectedly scarce. VanBurskirk and Simpson (2013) investigated the relationship between meteorological data (barometric pressure, humidity, outdoor temperature, and moon illumination) with classroom-collected behavioral data of children with autistic disorders [7]. Interestingly, in contrast, they found a weak relationship between the two [7]. However, due to the methodological limitations of the study, a further investigation should be conducted among children with complex neurological and/or motor impairments [7].

## 1.2. Children with PIMD/SMID

A condition that is characterized by profound intellectual disability (ID) (estimated intelligence quotient or IQ of less than 25) and restricted or absence of hand, arm, and leg functions is called profound intellectual and multiple disabilities (PIMD) or also known as severe motor and intellectual disabilities (SMID) [8-11]. One of the most challenging parts of supporting the children with this condition is communication, particularly understanding spoken or verbal language and symbolic interaction with objects [8,10]. Communication is limited to movements, sounds, body postures, muscle tensions, or facial expressions on a presymbolic (nonsymbolic) or protosymbolic level with no shared meaning and often minute and refined, which hinders expressing their needs [12-15]. In actual support settings, time has been used to interpret words and actions (patterns of movements and motions) of individuals with PIMD/SMID [16]. For example, in a nursing home station, an adult patient has been making sounds while flexing and extending his right leg. Looking at the time (11:40 am), the caregiver interprets that the movements and sounds mean that the patient is hungry and wants to eat. The caregiver confirms it by saying "It's lunchtime, and you're hungry, I'll prepare lunch for you." After that, the patient stopped making sounds and moving his right leg. These behavior and movements are unique and distinct in each child which most of the time, only their close caregivers (e.g. parents, teachers, therapists, etc.) can interpret, making it hard for others to be part of their communication group and require more support from their close caregivers thus, consequently decreases their independence and mobility [8,10, 17].

## 1.3. Attuning theory: Categorizing the behavior of children with PIMD/SMID

The possibility of categorizing the expressive behavior of children with PIMD/SMID was investigated by Ashida and Ishikura (2013) when they introduced six major categories based on the body parts movements involved in each expressive behavior of children with PIMD/SMID: eye movement, facial expression, vocalization, hand movement, body posture, body movement and non-communicative behaviors (others) [12]. More recently, Griffiths and Smith (2017) discussed in length how people with severe or PIMD communicate with others by introducing Attuning Theory [18]. This theory suggests that the communication between individuals with

PIMD/SMID and their caregivers is regulated by the process of attuning which describes how they move towards or away from each other cognitively and affectively [18]. This core category consists of seven discrete but dynamically interrelated categories namely setting, being, stimulus, attention, action (including maneuvering), engagement, and attuning [18]. They described and connected each category in a synopsis where all communication occurs in an environment or setting which described as the total context of a place (e.g. bus, field, kitchen, etc.) which influences the individuals' feelings and their state of mind (their being) or their action [18]. The being influences how each person behaves which is the stimulus that impacts the way people attend to each other (attention) and the nature of the interaction (engagement) [18]. Whether or not an action or engagement occurs is determined by the process of attuning which affects and reflects how the individuals perceive or feel their state of being and deliver stimulus to each other [18]. Further, they also comprehensively described in detail the structure (anti and pro and negative and positive), typologies (from screaming to harmony), indicators (looking at each other, the movement towards each other, smile, close physical contact, gaze, expression, etc.) and codes (concentration, interest, and support) of attuning [18]. They also listed and described the behavior and movement manifestations of each category (e.g. attention is manifested by visual tracking, mobile gaze changes, still gaze, head position, etc.) [18]. In conclusion, the proponents emphasized that the theory of attuning aids in predicting how and what to expect in communication with individuals with PIMD/SMID which also have important implications for practice whether it may be academic, clinical, or for leisure [19].

### 1.4. Friendly VOCA app, ChildSIDE app, and Script theory

Electroencephalography (EEG) in brain-computer interfaces (BCI), facial features extraction, eye tracking, and movement recognition, picture exchange communication, and app-based voice output communication aid (VOCA) are some of the most developed and advanced systems that aid in communication and the interpretation of the needs of children with PIMD/SMID [17]. Recently, aside from time, the importance of determining location has also been proposed to better understand the context of the interaction of individuals with PIMD/SMID with other people and the environment [16]. Friendly VOCA iOS mobile app has been developed based on the notion of Scripts Theory or the use of schema responding to or behaving appropriately to a particular situation or location [19]. Using map coordinates from Global Positioning System (GPS) and iBeacon, an indoor Bluetooth device that transmits short distance signals to the app, it automatically switches interfaces and displays depending on the user's location at a specific time [16]. Consequently, this idea has also motivated the development of ChildSIDE, an app that collects the interpretations of children's expressive behaviors with associated location and weather parameters for communication and independent mobility. Its accuracy in collecting and transmitting outdoor location using GPS, indoor location using iBeacon devices, weather parameters such as temperature and humidity, atmospheric pressure and wind direction and speed using ALPS Bluetooth sensors and OpenWeatherMap API has been tested which ranged from 82% to 93% [16].

### 1.5. Machine-learning-based behavior classification (classifiers, dataset recalibration, classes)

Children with autism spectrum disorder (ASD), attention-deficit and/or hyperactivity disorder (AD/HD), severe intellectual disabilities with cerebral palsy (CP), or other physical impairments have been significantly targeted by recent behavior studies that used machine learning [20-27]. Some of the most common and considered classification models are support vector machine (SVM), decision trees, random forest (RF), and neural network (NN) [20-27]. The variances in the classification accuracy rates between or among outcome classes were investigated in differentiating disorder sub-populations, distinguishing behavioral phenotypes between and among disorders, predicting or preliminary risk evaluation and/or pre-clinical screening and triage, diagnosis, and treatment, intervention, or rehabilitation [20-27]. The variances in the classification performance of different models, recalibrating dataset combination, and classifying multiclass behavior outcomes were also investigated and compared where SVM, recalibrated dataset combination, and 3 classes obtained higher accuracy rates respectively [20,27]. Further, sourced behavior data were from phenotypic, neuroimaging, electrodermal activity (EDA) or virtual reality (VR) systems and standardized scales and instruments or task or performance-based measures, yet, behavior data collected directly from observation of the behavior of the targeted children and their caregivers have not been utilized for behavior classification [20-26].

**1.6. Classification performance and feature selection method**

In non-human-behavior context, a proposed method in improving classification performance is the use of the Boruta feature selection method, especially when boosting the performance of the SVM classifier [28]. Further, a simulation study conducted to systematically evaluate and compare the performances of variable selection approaches revealed that Boruta was the most powerful approach compared to other variable selection approaches like Altman, Permutation approach, recurrent relative variable importance, or r2VIM, Recursive feature elimination, or RFE and Vita [29]. Boruta also showed stable sensitivity in detecting causal variables while controlling the number of false-positive findings at a reasonable level [29]. Kursa (2014) also identified Boruta as the most stable feature selection method in terms of the number of selected potentially important features when it was compared with other state-of-the-art RF-based selection methods [30].

Currently, there is no known investigation on whether environment data such as location and weather parameters can be used to classify the behavior or movement patterns using different machine learning models, nor there exists any behavior study that has sourced data from direct observations from children and their caregivers or any study that aimed to classify the behavior of children with cognitive and motor disorders, more so, of children with PIMD/SMID, for communication support and needs intervention. Therefore, this study aims to discriminate the behavior of children with PIMD/SMID to binary and multiclass behavior outcomes and investigate whether the inclusion of environment data would allow more accurate classification of major and minor behavior categories. The classification accuracy performances of different machine learning models in classifying different behavior outcome classes were also investigated and compared. We also explored whether training the dataset with a feature selection method would increase the classification accuracy rates. Lastly, the influences of and the interaction among behavior categories, with or without environment data, feature selection, and different machine learning classifiers in increasing classification accuracy rates were also evaluated and described. This study is exploratory in terms of investigating the use and inclusion of environment data in classifying the behavior of children and training the dataset with feature selection, however, we hypothesized that a high classification accuracy rate could be achieved in conducting such approaches. Based on previous studies, we also hypothesized that classification accuracy performance would be higher in classifying binary behavior outcomes and using an SVM-based classifier.

## 2. Methods
### 2.1. Participants, sessions, and experimental setup

Multiple single-subject face-to-face and video-recorded sessions were conducted among 20 purposively sampled children whose ages were from eight to 16 years old (3$^{rd}$ grade to 1st-year high school), who were mostly males (68%) and had either PIMD/SMID (n = 15; 79%) or severe or profound intellectual disabilities (n = 4; 21%) and their caregivers. The sessions were conducted based on international ethical guidelines and informed consent from all the primary caregivers of the children was obtained [31,32]. A total of 105 sessions were conducted with an average of five sessions per child and on average, ranged from a minimum of one session and a maximum of 15 sessions per child.

To capture the normal behavior of the children during and before the start of their classes, at lunchtime, and during break time and their interaction with their caregivers and other children, all sessions were recorded in the locations (e.g. classrooms, music rooms, etc.) where they usually spend time. Initially, to identify the behaviors under investigation, we used the categories and description of the body parts and movements involved in the target behaviors of children with PIMD/SMID from the study done by Ashida and Ishikura (2013) and by Griffiths and Smith (2017) among children (and their caregivers) with PIMD/SMID [12,18]. However, most of the time we had to add new ones as required especially when a child shows some reaction (e.g. vocalization, gesture) and when the caregiver responds by confirming the child's need (e.g. want to go to the toilet) verbally or by actions (e.g. assist the child to the toilet).

The average video recording time was 18.5 minutes (ranged from 0.37 to 54 minutes video recording time with a standard deviation or SD of 12.5 minutes) per session. We set up one videotape recorder (VTR) in a tripod 2 meters away from the subjects to capture the child's facial expressions and upper and lower limb movements and all the exchanges of responses between the children and their caregivers as shown in **Figure 1**. We also installed the weather and location sensors in each location and were placed either on a shelf, blackboard, bulletin board, or on an air-conditioning unit with approximately 2-meter distance from the investigator who used the ChildSIDE mobile app to input and transmits child's behavior, location, and environment data to the database.

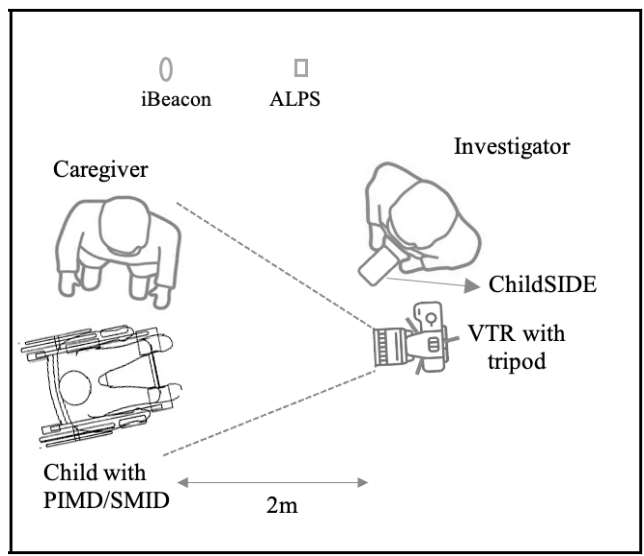

**Figure 1.** Intervention set-up. (a) Videotape recorder (VTR) focuses on the facial, upper, and lower limbs movements, (b) Intervention set up in a classroom setting: 2-meter distance from the VTR to the child with PIMD/SMID and caregiver, and the location where the sensors were placed.

### 2.2. ChildSIDE app and environment data sources

We used a previously developed app called ChildSIDE, an Android (7.0) mobile app to automatically record the caregivers' interpretation of children with PIMD/SMID's expressive behaviors and detect and transmit timestamps, outdoor and indoor location, and weather data to the database. The android's built-in time stamps and GPS (GPS/AGPS/Glonass) (a) was used to identify the user's current outdoor location in terms of map coordinates [33] while iBeacon (BVMCN1101AA B) from module BVMCN5103BK manufactured by Braveridge [34] was used to transmit indoor location data based on Bluetooth low energy (BLE) proximity sensing. ALPS Bluetooth sensors (The IoT Smart Module Sensor Network Module Evaluation kit), are multi-function Bluetooth sensors module (Mouser and manufacturer number: 688-UGWZ3AA001A Sensor Network Kit W/BLE Mod Sensors) developed by ALPS Alpine to acquire and transmit 11 motion and environment data using multiple sensors for pressure, temperature, humidity, UV, ambient light and 6-axis (Accel + Geomag) [35]. Other weather data (atmospheric pressure, humidity, sunrise, and sunset time) were obtained from OpenWeatherMap Application Programming Interface (API), an online service that provides weather data that matches the user's current location [36]. When a user clicks a behavior, the app automatically sends the behavior name with its associated location and weather data from the data sources to the Google Firebase database (**Figure 2**).

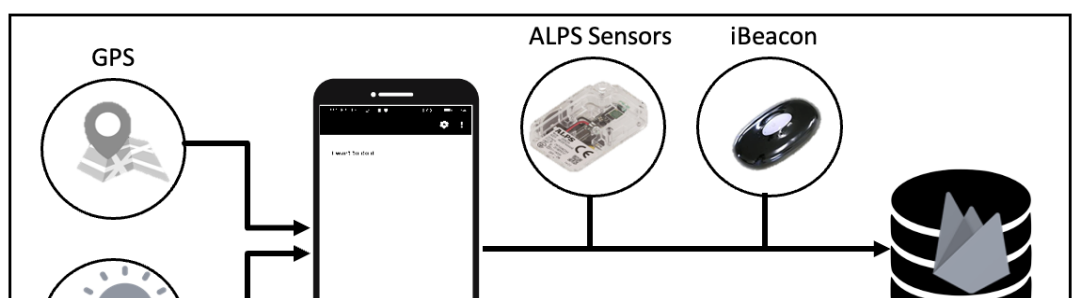

**Figure 2.** Data flow from the data sources (iBeacon, GPS, ALPS Sensors, and OpenWeatherMap API) detected and transmitted by ChildSIDE app to Google Firebase database.

**2.3. Data analyses workflow**

**Figure 3** illustrates the workflow of the data analyses that we conducted. Our datasets consisted of the characteristics (gender and condition) of the children with PIMD/SMID (CC), their behavior that was categorized into major (MajC) and minor (MinC) categories, and environment data (ED) that consist of location and weather data collected using the app from sensors and API. To investigate whether recalibrating the dataset with the inclusion of environment data would allow a more accurate classification of either major or minor categories of the behaviors, we made several combinations. Each dataset combination has child characteristics with major or minor (or both) behavior categories with or without environment data. The individual data that we collected were also categorized into 2, 3, and 7 behavior outcomes that formed the classes. The dataset combinations in each class were trained with feature selection (using Boruta) or without feature selection (non-Boruta) to investigate whether this method will improve the computational speed and accuracy of four classifiers or models: eXtreme Gradient Boosting (XGB), support vector machine (SVM), random forest (RF), and neural network (NN) which were primarily compared based on their accuracy rates. We tested the classification accuracy among 48 patterns in dataset combinations (6), with and without feature selection (2), and classifiers (4). This was done using R (version 4.0.3) free software programming language [37].

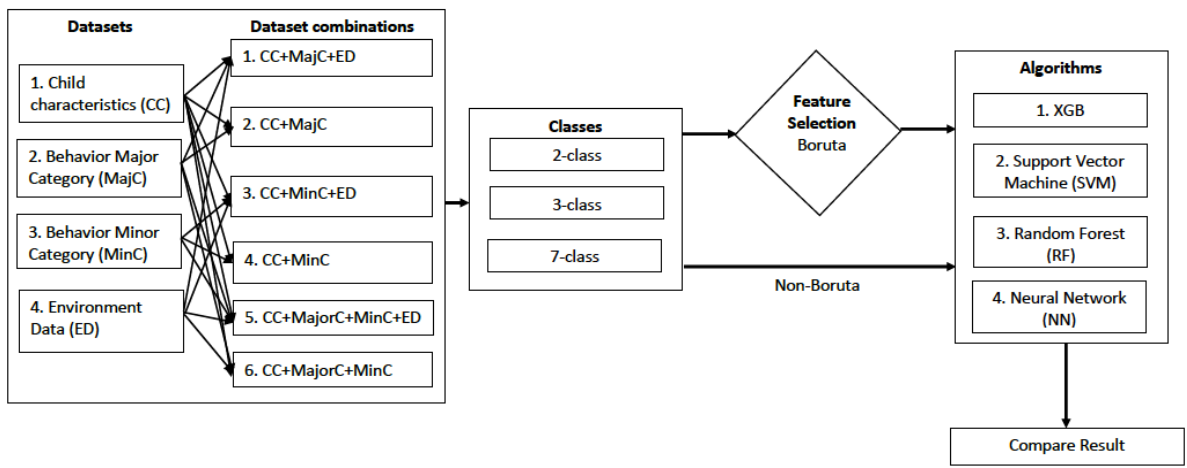

**Figure 3.** Data analyses workflow from dataset combination to classifier model accuracy comparison

**2.3.1. Child characteristics**

This dataset included children's gender (male or female) and their main condition as features. The children included in this study had main diagnoses of either profound intellectual and multiple disabilities (PIMD)/severe motor and intellectual disabilities (SMID) or severe or profound intellectual disabilities.

### 2.3.2. Major and minor behavior categories

Of the 292 individual behavior data that we collected, one had no score from the two raters, subjecting only the remaining 291 to inter-rater agreement Kappa statistics analysis. The video recordings were analyzed (at slow speeds frame-by-frame) by two raters to classify which body parts or movements were involved in each behavior. Each rater scored "1" in each minor category where a behavior belongs, otherwise, they scored it "0". Similarly, each major category was scored "1" where it had at least one minor category with a score of at least one. Then Kappa statistics were computed to identify the level of agreement between the two raters in each major and minor category. With fair (0.21-0.40) to almost perfect (0.81-0.99) agreement levels, the two raters reached a consensus and a final categorization of behaviors was created. There were 676 body parts or movements categorized into six major and 16 minor behavior categories. Vocalization was considered as a major and as a minor category. The description of each minor behavior or feature is shown in **Table 1**. Pre-processing for machine learning involved the conversion of major and minor categories to numerical data using integer and one-hot encoding.

### 2.3.3. Environment data (location, time, and weather) and k-NN imputation

GPS (GPS/AGPS/Glonass) that was used to identify the user's current outdoor location in terms of map coordinates has two features, 8-digit latitude and 10-digit longitude in decimal degrees (DD, e.g. 33.795098, 132.8702426). iBeacon, which was used to identify the indoor location, transmits proximity measurements based on three features: Radio Signal Strength Indication (RSSI), MAC address (6 bytes: F5:B0:E2:A2:AE:69), and iBeacon name to a close mobile device to identify the user's specific indoor location. ALPS multiple sensors (pressure, temperature, humidity, UV, ambient light and 6-axis (Accel + Geomag) was used to acquire and transmit 11 motion and environment data features: UV(mW/cm2) range (UV-A: 0 to 20.48; Ambient: 0 to 81900), ambient light (Lx) resolution (UV-A: 0.005; Ambient: 20), Accel+Geomag-g1, g2, g3 ranges (-2.4 to +2.4; Acc: -2 to +2), Accel+Geomag- µT1, µT2, µT3 resolutions (0.15; Acc: 0.24), atmospheric pressure(hPa) (300 to1100), temperature and humidity sensor range [℃] (Temp.: -20 to +60; Hum.: 0 to100), and temperature and humidity sensor resolution [%RH] (Temp.:0.02; Hum.: 0.016). Lastly, OpenWeatherMap API has 13 weather parameters or features: weather (e.g. cloudy, raining, etc.), sunset time, sunrise time, current time, minimum temperature (℃), maximum temperature (℃), atmospheric pressure (hPa), main temperature (℃), humidity (%), weather description, cloudiness (%), wind direction (degrees), and wind speed (meter/sec.). We have also included timestamps that consisted of seasons (order of months corresponds to the number that covers each season, e.g. 3 to 5 for spring, 6 to 8 for summer, etc.), date (year, month, day), and time (hours, minutes, and seconds) in the environment data.

Of the 364 collected individual behavior data, 35 individual behavior data that were not detected by the app and 37 without any associated data from any data source were deleted. In total, 292 individual behavior data had associated data from iBeacon, GPS, ALPS, Sensor, or OpenWeatherMap API data sources were used for the analyses. In terms of location data, all of the coordinates from GPS were detected and transmitted to the database (100%), however, there were only 246 (84.2%) data from iBeacon (MAC address, RSSI, and iBeacon names). Among the ALPS sensors data, there were 98 (33.6%) missing data in UV range, 15 (5.14%) in 6-axis (Accel+Geomag) sensor ranges, 14 (4.79%) in 6-axis (Accel+Geomag) sensor resolutions [µT], and 53 (18.1%) in UV resolution. Pressure sensor range [hPa] and temperature and humidity sensor range and resolution [%RH] had no missing data. OpenWeatherMap API had 14 (4.79%) missing data, except for wind direction which had 35 (11.9%) missing data.

In cases where there was only a single missing value in a dataset collected in a specific location and time within a session, the values of the nearest non-missing data were used to identify the possible value of that missing data. For example, a missing RSSI data of the iBeacon should be similar to the other non-missing RSSI data collected within the same session at the same time frame and location. However, in cases where there were multiple missing data and identifying which nearest non-missing data to input in the missing data was not possible, our preprocessing involved *k*-Nearest Neighbor (*k*-NN) imputation (*k* = 14). This was done to avoid relatively large average errors due to the anticipated decrease in the size of our dataset after deleting individual behavior data with no matched environment data [38]. *k*-NN impute is the KNN algorithm's function to impute the missing data values, which, compared to deleting the rows with missing data, can easily and effectively handle several missing values in large datasets [38]. This technique identifies the partial feature in a dataset and selects the K nearest data (by calculating the distance) from training data with the known values to be imputed. Then it substitutes the missing value with an estimated value (using the mean) of the data with known or non-missing data values [38]. Pre-processing involved transforming categorical data (e.g. iBeacon name, four seasons, days of the week, years, and months) to numerical data using integer and one-hot encoding.

**Table 1.** Feature description of major and minor expressive behavior categories

| Major and minor behavior categories | Feature description |
|---|---|
| **a. Eye movement** | |
| 1. Gazing | Gaze at people and things (look at the faces of individuals who are not in their communication group) |
| 2. Eye-tracking | Eye movements that follow the movements of people and things in a linear fashion |
| 3. Changing line of sight | Change of line of sight, movement of the line of sight; gaze rolls and moves; a point-like movement that is not "a.2. eye tracking." The momentary glare can also be evaluated. Movements that cannot be evaluated as gaze/tracking. |
| 4. Opening or closing the eyelids | Not an involuntary blink. Their reaction when told to open or close their eyes. |
| **b. Facial Expression** | |
| 1. Smiling | Smile |
| 2. Facial expression (other than a smile) | Something that is not expressionless. Changes in facial expressions. Surprise, frowning, sticking out tongue, etc. |
| 3. Concentrating and listening | Focusing on picture books, music, and voices, etc. |
| **c. Vocalization** | Producing sound |
| **d. Hand movement** | |
| 1. Pointing | Hand pointing or pointing finger towards an object. |
| 2. Reaching | The action of reaching or chasing after reaching the target, not by pointing hand or finger. |
| 3. Moving | Grab, hit, beckon, push, raise hands, dispel, etc. |
| **e. Body movement** | |
| 1. Approaching | The Head or upper body (or the whole body) is brought close to a person or an object. |
| 2. Contacting | Touching people and things with hands and body. It does not include cases that are touched by accident or touched. |
| 3. Movement of a part of the body | Head and neck movements, upper body movements, upper and lower limb movements (shake, bend, move the mouth, flutter legs, etc.); (excluding "d.1. pointing", "d.2. reaching", "d.3. moving"), etc. Distinguish from "f.1. stereotyped behavior" |
| **f. Non-Communicative Behaviors (Others)** | |
| 1. Stereotypical behavior | The same behavior or movement is repeated without purpose. Behavior that occurs in a certain repetition e.g. Finger sucking, shaking hands, rocking, etc. (Shaking things is "d.3. moving") |
| 2. Self- and others-injurious behavior | Hitting someone, biting a finger, etc. |

### 2.3.4. Binary and multiclass behavior outcome classes (Attuning theory)

The 292 videotaped individual behavior data were also subjected to inter-rater agreement Kappa statistics analysis by five independent behavior expert raters to interpret and classify which behavior outcome each behavior belongs to. Each behavior data was analyzed and coded that corresponds to an outcome. Initially, we grouped the behavior data into 9 behavior outcomes (first outcome level): calling, response, emotions, interest, negative, selecting, physiological response, positive and unknown interpretations (**Table 2**). The behavior outcomes that we created were similar to that of Attuning Theory's indicators, codes or categories like engagement (joint attention), assent, harmony, delight, please or pleasure, interest, and pro and negative attuning (refusal). However, we also developed new ones as required (selecting and physiological response categories). From the 9 behavior outcome categories, we had to delete the "unknown" and "positive" categories due to the small sample size (n = <4) thus, identifying operational definition from the extracted behavior manifestations was not possible.

**Table 2.** Feature description and behavior manifestations of the 7 behavior outcomes (class) in comparison with the Attuning Theory

| This study | | | Attuning Theory | |
|---|---|---|---|---|
| Category | Definition | Manifestations (sample extracts) | Category | Definition |
| Calling | Verbal (e.g. greetings, vocalization) or non-verbal behavior (e.g. smile, staring, pointing, etc.) aim to get the attention of the caregiver or teacher | -moves mouth to say only the "mas" part of "Ohayo gozaimasu"; -Move the face widely. Open mouth wide and try to speak. Breathe a little harder; -Vocalizes while touching the back of the neck with the left hand; -Looks out of the window at the car and says "Densha" (streetcar); -Pats the teacher on the back. Turning to face her and mumbling something. | Engagement (joint attention) | The engagement of both partners in the dyad may be directed to the same focus. |
| Response | Verbal (e.g. "yes", "bye-bye", etc.) or non-verbal (e.g. raises a hand, nodding, wave hands, clapping, etc.) responses to other's questions or gives signals to another person | -Pointing or pushing somewhere in the book with the left hand; -Looks at the teacher's face and says "yes". Makes a slight nodding movement; -Raises both arms upwards. Raising the corners of the mouth. Saying "mmm". Shake your head from side to side. Shake head vertically. Movement of the mouth. -Vocalization. Moving the body. Increased breathing. Eye movement. | Assent | demonstrates attuned agreement between the dyad. One partner acts or asks a question and the other responds in a clear affirmative manner. |
| Emotions | Mostly non-verbal expressions of feelings of being happy, pleasure, excited, perception of fun, angry, worried, troubled, etc. (e.g. smile, moving or opening mouth, shaking head vertically or body, looking away, etc.) | -being delighted, raises the corner of his mouth and shakes his face from side to side while holding the back of his head with his right hand; -Pleasant feeling, open mouth and bring the hand to mouth. Looks at the teacher and opens their eyes. Raises eyebrows upwards; -feeling angry, the body begins to sway. The corners of the mouth and the corners of the eyebrows fall. Suddenly stands up and walks over to the TV; -feeling troubled, move the right hand up and down in front of the face. Say the words Move the right hand back | Harmony, delight, pleased, pleasure | *Harmony,* characterized by actions and communications that display mutual satisfaction. *Pleased* demonstrates with smiles, grins, and other visual appearances of satisfaction, quiet contentment with what is going on. This differs from *pleasure,* describing a more intense satisfaction, where smiling veers towards laughter, where the communication is more intense and direct. |
| | | and forth. Touching the front teeth with the left hand. | | |
| Interest | Verbal (e.g. "let me see", "yes!", "what's that?") or non-verbal (e.e.g pointing, raising hands, standing up, nodding, etc.) that hints interest in an object, person, or action or doing an action. | -Opens mouth wide, smiles, and bends over. Says "Oh, hi, hi, hi" in a strained voice;<br>-Stares at a ball. Moves towards the toy;<br>-Pointing and saying "Oh, what?";<br>-Stands up and walks to the teacher in front of him;<br>-Says "woo". Touches front teeth with the left hand. Moves left hand to right ear Eye movement. Mouth movement Repeated raising and lowering of the right hand (in front of the face). Looks down;<br>-Smiling and nodding. | Interest | the communication partner demonstrates obvious attention and interest in (attuning to) the action that is going on. The attention is focused on the action. The result of the interplay of attention and action is that the attuning level of the partners rises and falls in tandem with the attention displayed to the action. |
| Negative | Verbal (e.g. "no", "don't like", "dislike" or "end") or non-verbal actions and vocalizations (e.g. closes mouth, sticks out tongue, turns face away) to express refusal or disagreement. | -refuses to take a spoonful of rice in his mouth. Closes his mouth when a spoon is put close to his mouth;<br>-pushes away the teacher's hand. Closes mouth. Slaps own body with the hand;<br>-frowns and pushes the teacher's body with his right hand;<br>-touches his face (mouth and nose) with his hands while moving his fingers. | Pro and negative attuning (Refusal) | In this state, pro-attuning coexists with negative attuning. The communication partners understand each other very well (high pro-attuning). However, they do not accede to the wishes of the other so the interplay between the dyad is negative. |
| Selecting | Mostly non-verbal actions or gestures (e.g. pointing, tapping, reaching) to express decision or desire to choose between or among objects. | -Points to a picture book. Says a sound similar to "this";<br>-Looks around at the side dishes and selects a side dish by saying "this one" with the index finger of the left hand;<br>-Looks at what the teacher is pointing at. Moves left hand;<br>-Tapping the teacher's/caregiver's foot;<br>-Points to the numbers on the board with hand. | - | - |
| Physiological response | Verbal (e.g. saying "rice", "sleepy", "thirsty", etc.) and non-verbal (e.g. closing eyes, not opening mouth) vocalizations and actions to express functions or desires relating to normal physical or bodily responses. | -sleepy, eyelids close. Look up and do not move;<br>-sleepy, mouth opening is too small. Refuses to take food in the mouth;<br>-thirsty, calls for the caregiver/teacher three times.<br>-sleepy, look down. Movement becomes stiff;<br>-in pain, frowning and touching the teacher's hand with the right hand;<br>-tired, plops down on the desk. Sneezing.<br>-says "rice" while looking at the table. | - | - |

The individual behavior data were then analyzed and grouped into 3 behavior outcomes (2nd outcome level) which are response, action, and response or action (**Table 3**), similar in partial with Attuning theory's stimulus (non-action), action (dual response) and the relationship between stimulus and action category definitions [18]. In the last outcome level, the third behavior outcome, the behavior, movements, gestures, facial expressions, vocalization, or other behavior of children with PIMD/SMID have categorized either response or action using the definition used in the 2nd outcome level (**Figure 4**). In each outcome level, Kappa statistics were computed to identify the level of agreement between and among the experts. In cases when there were low agreement levels (1.01 to 0.60), a series of pair (first and second outcome levels) and group (third outcome level) expert brainstorming sessions were conducted until an acceptable to almost perfect agreement levels (0.61 to 0.99) were reached before proceeding to another outcome level.

### 2.3.5. Feature selection (Boruta)

Boruta was used for feature selection as it provides an unbiased selection of important features and unimportant features from an information system [30]. It identifies important features useful for model prediction from the randomly expanded system produced by merging the randomly permuted features and the original training features [30]. To assess the importance of the feature in the original system, Z-scores of the original features and the randomly permuted features were compared, in which the maximum Z-scores among the randomly permuted features were identified [30]. If the Z-score of the feature was higher than that of the maximum Z-scores among the randomly permuted features, the feature was deemed important, otherwise, not important. For each run, the randomly permuted data was different. Using maximal importance of the randomly permuted features, a two-sided binomial test identifies an important feature if the number of times it was found important (observed) was significantly higher than the expected number (0.5N) at significance level a, otherwise, will be excluded from the system in subsequent runs [30]. We implemented it in the R package Boruta which uses an RF method using the maxRuns parameter which will force the run to stop prematurely. The features that neither important nor unimportant were tagged as tentative. In this study, as recommended by default, we set the confidence level a = 0.01 and run the algorithm with 100 maxRuns due to fewer features were included in the model.

**Table 3.** Feature (category) description of the 3 and 2 behavior outcomes (class) in comparison with the Attuning Theory

| This study | | | | Attuning Theory | |
|---|---|---|---|---|---|
| Outcomes | | Category | Definition | Category | Definition |
| 3 behavior outcomes | 2 behavior outcomes | Response | a one-way communication (from the perspective of the caregiver/teacher) that stimulus from the child (movements, gestures, facial expressions, vocalization, or other behavior) may affect or influence the attention of the caregiver or teacher but don't necessarily require an active response from the caregiver/teacher. | Stimulus (Non-action) | Stimulus is an attempt by one partner to encourage action from another partner.<br>Non-action is concerned with settings where minimal stimuli are present, but no action is elicited from the participant (s). It may be passive or active (determined inaction) or occur as a result of a period of stasis. |
| | | Action | two-way or mutual communication (from the perspective of the caregiver/teacher) where the stimulus from the child (movements, gestures, facial expressions, vocalization, or other behavior) affects or influence the attention of the caregiver or teacher which cause a response through action (e.g. attending to children's needs). | Action (Dual action) | Actions are an observable process of behavioral change in an individual that is demonstrated by movement, gestures, facial expression, vocalization, or other behaviors. It can be a dual-action where action may be carried out by both participants in the dyad, that is, they may work together to achieve an action. Dual-action arises where one participant carries out part of an action, but the other completes it. |
| | Response/Action | | stimulus from the child (verbal or non-verbal responses or behavior manifestations through movements, gestures, facial expressions, vocalization, or other behavior) which affect or influence the attention of the caregiver or teacher which may or may not require responding through action. | | *"The dividing line between these concepts is that a code is grouped under action if it comes about as a result of a previous stimulus or is to be an event that is not designed to elicit a reaction, whereas a code is grouped under stimulus if it clear that a stimulus is provided to induce a certain course of action."* |

### 2.3.6. Classification methods (classifiers)

We tested the classification performances of eXtreme gradient boosting (XGB), support vector machine (SVM), random forest (RF), and neural network (NN) predictive models on our dataset combinations (major and minor behavior categories with and without environment data) trained with or without feature selection (Boruta) to conduct binary and multiclass classes. Dataset has been partitioned into two parts (training and testing) where 80% were validation data which were tested with 20% remaining data to tune for hyperparameters. A stratified extraction method was used to avoid bias in dividing the data. To avoid the problem of overfitting and underfitting, 10-fold cross-validation was done. The selection of the classification models included in this investigation as per the most common and considered models from previous behavior studies particularly SVM, RF, and NN. Considered as a powerful method for classification which creates hyperplane (maximum of p-1 planes) between classes (datasets with same properties considered as p-dimensional vectors) to predict labels from support vectors, SVM attempts to identify the best classifier/hyperplane which is located in the middle which has the maximum margin between the two classes (maximum-margin hyper-plane). Another algorithm that was designed to build several decision or classification trees that are consolidated to produce a precise and stable prediction is RF. In this model, each tree is constructed using different "bootstrap" or the same number of randomly selected samples from the original sample data used as its replacement. From this random sample, an estimated 63% of the original sample occur at least once and the remaining 1/3 was not used to build a tree and instead used for performing out-of-bag error estimate and feature importance. Then it randomly selects data nodes to construct a decision tree. Another predictive model, NN is consist of nodes of artificial neurons which are represented by some state (0 or 1) or weight assigned to them to define its importance in each layer, from input, middle (hidden), an output layer, and the desired outcome that is more predictive of the specified outcome. The number of classes in multiclass classification commonly corresponds to the number of nodes in the outer layer. In addition, we also included a model that has robust power to control overfitting which extends the Boosting Tree models by performing second-order Taylor expansion of the objective functions after each split and adds splitting threshold (except if the information gain is greater than the threshold). The classification performance of all four models was compared by accuracy, precision, recall or sensitivity, specificity, and the area under the curve (AUC). Accuracy indicates how the classifier is often correct in the diagnosis of whether the major or minor categories are better with environmental data or not, while precision has been used to determine the classifier's ability to provide correct positive predictions of the behavior. Recall or sensitivity and specificity were used to identify the proportion of actual behavior correctly identified by the classifier and to determine the classifier's capability of determining negative cases of behavior, respectively. The average AUC was also used to assess the goodness of a classifier's prediction which resulted from the 10 cross-validation trials where an AUC value near 1 is termed as the optimal classifier.

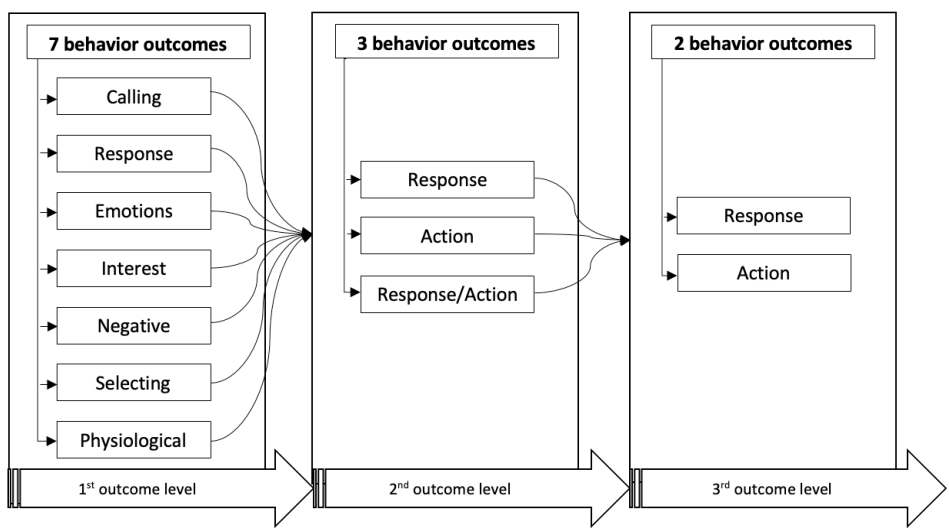

**Figure 4.** Features (categories) in each behavior outcomes (classes) and levels

## 2.4. Statistical analysis

Pre-comparison, the classification accuracy rates (%) of the 10-fold cross-validation results were averaged to obtain the mean classification accuracy rates (%). We conducted one-way ANOVA (2-tailed) with Bonferroni *posthoc* test to identify the dataset combination with the highest accuracy rate (%) among the combinations and within classes. We used the univariate General Linear Model (GLM) feature of SPSS to conduct two-stage three-way ANOVA to analyze the influences of and evaluate the interaction among the classes, dataset (with and without environment data), feature classification (with and without Boruta), and the machine learning algorithms to the classification accuracy rates using Bonferroni *posthoc* test (influences and interactions with a significance level of $P<0.05$). The mean classification accuracy rates (continuous) were pooled and entered as the dependent variable while the factors were coded as categorical independent variables: dataset (with the environment "1", without environment "0"), feature classification (with Boruta "1", without Boruta "0"), and algorithm (XGB: 1, SVM: 2, RF: 3, NN: 4), and class (class 2 coded "1", class 3 coded "2" and class 7 coded "3"). The degree of the influences of and the interactions among the factors to the mean accuracy was identified using the partial eta squared ($\eta^2$) and a subsequent Bonferroni *posthoc* test determined the differences between the mean accuracy rates of the factors and interactions with significant variances ($P <.05$).

## 3. Results
### 3.1. Among recalibrated dataset combination

The highest classification accuracy rates were found in the Boruta-trained CC+MinC+ED dataset using the NN algorithm in class 2 and non-Boruta CC+MinC+ED dataset using SVM classifier in class 3 (M = 76%). After combining the classification accuracy rates of all the algorithms in each dataset combination, Boruta-trained CC+MinC+ED in class 2, non-Boruta CC+MinC+ED, and non-Boruta CC+MajC+MinC+ED datasets obtained the highest mean classification accuracy rate of 73% (**Supplementary File S1**).

We then averaged the classification accuracy rate in each dataset combination per class and we found that in class 2, the datasets with the significantly highest mean classification accuracy rates were CC+MajC+ED (a) (70%), CC+MinC+ED (c) (72%), and CC+MajC+MinC+ED (e) (72%) ($P<.001$). All dataset combinations in class 3 have the significantly highest mean classification accuracy rates except CC+MajC (b) (63%) and CC+MajC+MinC (f) (67%). In class 7, dataset combinations a, c and e, had the highest classification accuracy rates of 45%, 46%, and 46%, respectively. There were significant differences among the mean classification accuracy rates among the dataset combinations within class 2 ($P<.001$), class 3 ($P<.001$), and class 7 ($P<.001$) as shown in **Figure 5**. When we compared the dataset combinations with the highest accuracy rates between classes, one-way ANOVA revealed that datasets a, c, and e of classes 2 and 3 had the highest accuracy rates among all the dataset combinations ($P<.001$ *posthoc* Bonferroni test).

### 3.2. Within classes

We then averaged the classification accuracy rate by dataset (with and without environment data) in each class. In classifying 2 behavior outcomes (response versus action), we obtained an average accuracy of 66.8% (SD = 1.59) as shown in **Table 4**. The highest and lowest mean classification accuracy rates by dataset were obtained in Boruta-trained dataset with environment data using NN classifier (74.7%) and Boruta-trained dataset without environment data using XGB and SVM classifiers (59%), respectively. Further, datasets with environment data (either with or without Boruta feature selection) (71.6%) were significantly higher than datasets without environment data (62%, *P*<.001). The confusion matrix showed that the response and action behavior outcomes were classified correctly with lower confusion in binary class (**Figure 6**).

The mean classification accuracy rate for classifying 3 behavior outcomes (response, action, and response or action) was 68.6% (SD = 0.70). The mean classification accuracy rates ranged from 65.4% to 73.5% where the lowest was obtained by the non-Boruta-trained dataset without environment data while the highest was obtained by the non-Boruta-trained dataset with environment data (Table 4 in the middle). Both datasets were ran using the SVM machine learning algorithm. SVM and RF classifier-based datasets had higher mean classification accuracy rates of 69.2% than NN (68.5%, SD = 2.43) and XGB (67.7%, SD = 2.00) classifiers. **Table 5** shows that

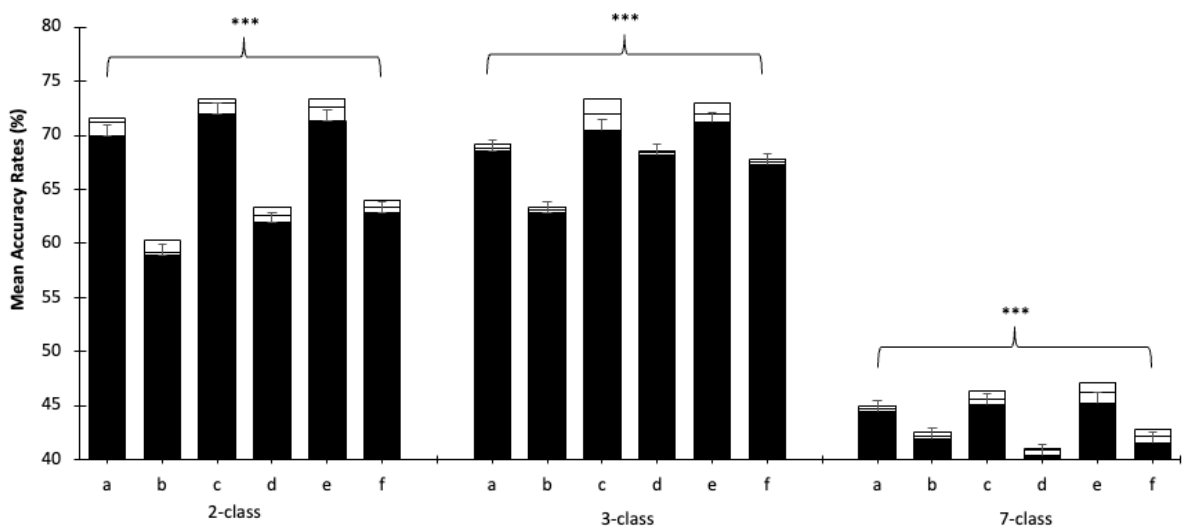

**Figure 5.** Mean classification accuracy rates (%) of each dataset combination (a-f) within the class (class 2, class 3, and class 7) and between class (similar dataset combination were compared across the classes) comparison using one-way ANOVA (2-tailed) with Bonferroni *posthoc* test.

Note: a = child characteristics with major behavior category and environment data; b = child characteristics and major behavior category; c = child characteristics with minor behavior category and environment data; d = child characteristics and minor behavior category; e = child characteristics with major and minor behavior categories and environment data; f = child characteristics with major and minor behavior categories; *P*<.001***, *P*<.01**, *P*<.05*.

RF and NN had the highest specificity of 58%. These two classifiers together with SVM had the highest recall/sensitivity of 81%. The highest AUCs were obtained by SVM and NN (74%) while the highest precision and F1 score were obtained by SVM and RF (68%). Similar to 2-class, the datasets with environment data (either non- or Boruta-trained) (70.9%) were significantly higher than datasets without environment data (66.3%, *P*<.001). The confusion matrix shows that in contrast with the response and response or action behavior outcomes, classifiers had difficulty classifying action behaviors that were incorrectly classified as response behavior outcomes (**Figure 6**).

In classifying 7 behavior outcomes (calling, emotion, interest, negative, physical response, response, and selection), the mean classification accuracy rate was 43.6% (SD = 0.55). The highest mean classification accuracy rate of 47.1% was obtained by the non-Boruta dataset with environment data in the RF algorithm while the lowest was obtained by the Boruta-trained dataset without environment data in the SVM classifier (38.6%).

RF classifier-based dataset had higher mean classification accuracy rates of 44.4% than the other classifiers (43.3%, SD range = 1.47 to 3.45). Surprisingly, all the classifiers had 90% specificity in classifying 7 behavior outcomes, the highest among the outcome classes (**Table 4**). Consistently, datasets with environment data (either non- or Boruta-trained) (45.6%) were significantly higher than datasets without environment data (41.6%, *P*<.001). The classifiers had incorrectly classified physiological response as response and selection as interest behavior outcome **(Figure 6)**.

**Table 4.** The classification accuracy rate by dataset (with and without environment data) in each class.

| | 2-class | | | | | 3-class | | | | | 7-class | | | | |
|---|---|---|---|---|---|---|---|---|---|---|---|---|---|---|---|
| | (+) Env | | (-) Env | | Acc. | (+) Env | | (-) Env | | Acc. | (+) Env | | (-) Env | | Acc. |
| | (+)Bor | (-)Bor | (+)Bor | (-)Bor | Mean (SD) | (+)Bor | (-)Bor | (+)Bor | (-)Bor | Mean (SD) | (+)Bor | (-)Bor | (+)Bor | (-)Bor | Mean (SD) |
| XGB | 69.0 | 67.6 | 59.1 | 64.4 | 65.0 (4.39) | 69.3 | 69.6 | 65.8 | 66.2 | 67.7 (2.00) | 43.8 | 46.7 | 39.6 | 43.2 | 43.3 (2.92) |
| SVM | 72.1 | 72.0 | 59.1 | 62.9 | 66.5 (6.57) | 70.7 | 73.5 | 65.4 | 67.2 | 69.2 (3.61) | 46.5 | 45.1 | 38.6 | 43.1 | 43.3 (3.45) |
| RF | 71.6 | 73.3 | 59.5 | 63.4 | 66.9 (6.58) | 70.9 | 73.3 | 65.6 | 66.9 | 69.2 (3.56) | 46.6 | 47.1 | 41.2 | 42.8 | 44.4 (2.88) |
| NN | 74.7 | 72.6 | 61.9 | 66.3 | 68.9 (5.86) | 69.6 | 70.9 | 65.5 | 67.8 | 68.5 (2.34) | 44.5 | 44.3 | 41.3 | 43.1 | 43.3 (1.47) |
| Mean (SD) | 71.85 (2.34) | 71.4 (2.57) | 59.9 (1.35) | 64.2 (1.50) | 66.8 (1.59) | 70.1 (0.79) | 71.8 (1.90) | 65.6 (0.17) | 67.0 (0.67) | 68.6 (0.70) | 45.4 (1.42) | 45.8 (1.32) | 40.2 (1.31) | 43.1 (0.17) | 43.6 (0.55) |

Note: (+) Env = dataset with environment data; (-) Env = dataset with environment data; (+) Bor = with Boruta feature selection; (-)Bor = without Boruta feature selection; XGB = eXtreme Gradient Boosting; SVM = support vector machine; RF = random forest; NN = neural network; SD = standard deviation

**Table 5.** Classification performance rates of the classifiers in each class.

| | 2-class | | | | | 3-class | | | | | 7-class | | | | |
|---|---|---|---|---|---|---|---|---|---|---|---|---|---|---|---|
| | Rec. | Spec. | Prec. | F1 | AUC | Rec. | Spec. | Prec. | F1 | AUC | Rec. | Spec. | Prec. | F1 | AUC |
| XGB | 0.70 | 0.59 | 0.68 | 0.68 | 0.67 | 0.55 | 0.80 | 0.67 | 0.67 | 0.72 | 0.39 | 0.90 | 0.46 | 0.44 | 0.71 |
| SVM | 0.72 | 0.60 | 0.70 | 0.70 | 0.67 | 0.57 | 0.81 | 0.68 | 0.68 | 0.74 | 0.39 | 0.90 | 0.46 | 0.44 | 0.67 |
| RF | 0.69 | 0.64 | 0.71 | 0.69 | 0.68 | 0.58 | 0.81 | 0.68 | 0.68 | 0.73 | 0.40 | 0.90 | 0.48 | 0.45 | 0.72 |
| NN | 0.70 | 0.67 | 0.73 | 0.70 | 0.70 | 0.58 | 0.81 | 0.67 | 0.67 | 0.74 | 0.39 | 0.90 | 0.45 | 0.43 | 0.70 |

Note: XGB = eXtreme Gradient Boosting; SVM = support vector machine; RF = random forest; NN = neural network; Rec. = recall; Spec. = specificity; Prec. = precision; F1 = F1 score; AUC = area under the ROC curve

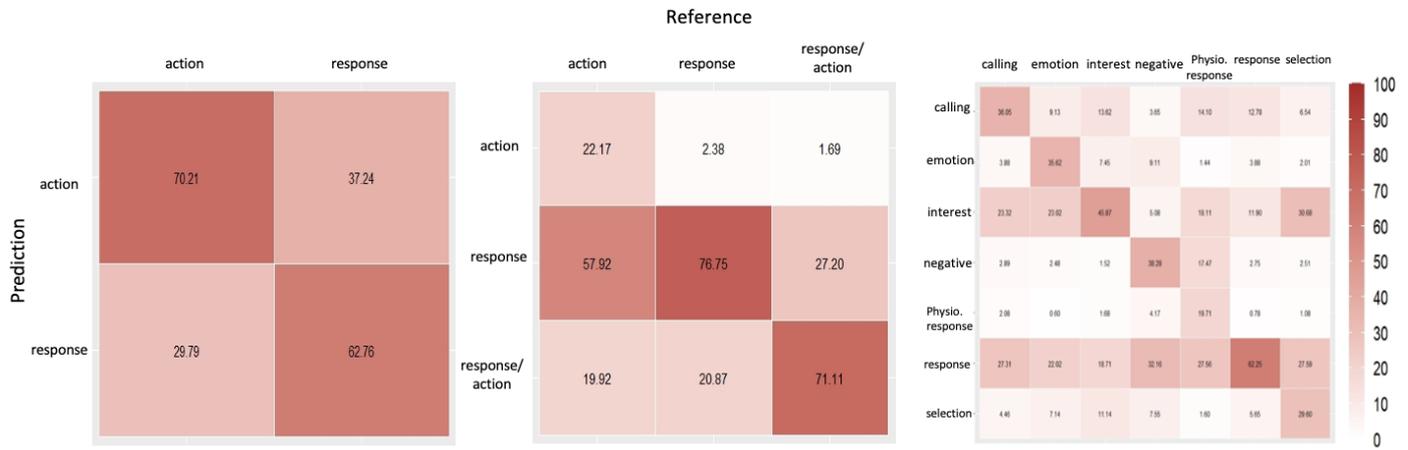

**Figure 6.** confusion matrices showing the mean classification accuracy (%) for all dataset, feature selection, and algorithms combination within each class

### 3.3. Factor influences to the overall mean classification accuracy rate

After pooling the mean accuracy rates and adding the classes as an independent variable, three-way ANOVA found that dataset, feature selection, classifier, and class had significant effects on the mean classification accuracy rates ($P<.001$) (**Table 6**). The high variances of 98% and 79% in the mean classification accuracy rates were attributed to dataset and class, respectively. Dataset with environment data (63%) (Figure 7a), non-Boruta trained (61%) (**Figure 7b**), and class 3 (69%) (**Figure 7d**) had significantly higher mean classification accuracy rates than dataset without environment (57%), Boruta-trained (59%), and classes 2 (67%) and 7 (44%) ($P<.001$), respectively. SVM, RF, and NN classifiers (60%) (Figure 5c) had significantly higher mean classification accuracy rates than the XGB classifier (59%). In particular, SVM and RF had the highest recall/sensitivity of 56% while RF and NN had the highest specificity (79% and 80%, respectively), precision (62%), and AUC (71%). RF also had the highest F1 score of 61% among the classifiers.

**Table 6** also shows that the interactions between dataset with feature selection, between dataset with algorithms, and between dataset with classes were found to have significant effects on the mean classification accuracy rates ($P<.001$, $P = 0.004$, and $P<.001$, respectively). Non-Boruta trained dataset with environment data (63%) (**Figure 7e**), a dataset with environment data using RF classifier (64%) (**Figure 7f**), and a dataset with environment data in class 2 (72%) (**Figure 7g**) had significantly higher mean classification accuracy rates. The interaction between classifier and class ($P = 0.01$) and the interactions among dataset, feature selection, and class ($P = 0.007$) also had significant effects on the mean classification accuracy rates. The use of NN in class 2 (69%), SVM (69%), and RF (69%) in class 3 (**Figure 7h**) and Boruta-trained dataset with environment data in class 2 (72%) and non-Boruta dataset with environment data in class 3 (72%) (**Figure 7i**) also had significantly higher mean classification accuracy rates.

**Table 6.** Three-way ANOVA results of the influences of dataset, feature selection, classifier, and class (and interaction factors) to the mean classification accuracy rates (%)

| Factors | df | F-value | η² |
| --- | --- | --- | --- |
| Dataset | 1 | 351.79*** | 0.79 |
| Feature Selection | 1 | 28.29*** | 0.23 |
| Classifier | 3 | 4.77*** | 0.13 |
| Class | 2 | 2491.16*** | 0.98 |
| Dataset * Feature selection | 1 | 12.96** | 0.12 |
| Dataset * Classifier | 3 | 4.64*** | 0.13 |
| Dataset * Class | 2 | 29.77*** | 0.38 |
| Feature selection * Classifier | 3 | 0.23 | 0.01 |
| Feature selection * Class | 2 | 0.10 | 0.00 |
| Classifier * Class | 6 | 2.88* | 0.15 |
| Dataset * Feature selection * Classifier | 3 | 0.75 | 0.02 |
| Dataset * Feature selection * Class | 2 | 5.20** | 0.10 |
| Dataset * Classifier * Class | 6 | 0.82 | 0.05 |
| Feature selection * Classifier * Class | 6 | 0.82 | 0.05 |
| Dataset * Feature selection * Classifier * Class | 6 | 0.81 | 0.05 |
| Error | 96 | | |

Note: Dataset = child characteristics, major and minor behavior categories with or without environment data; feature selection = with or without Boruta algorithm; classifier = XGB, SVM, RF or NN; class = class 2, 3 or 7; η² = partial eta squared; adjustment for multiple comparison (Bonferroni); P<.001***, P<.01**, P<.05*

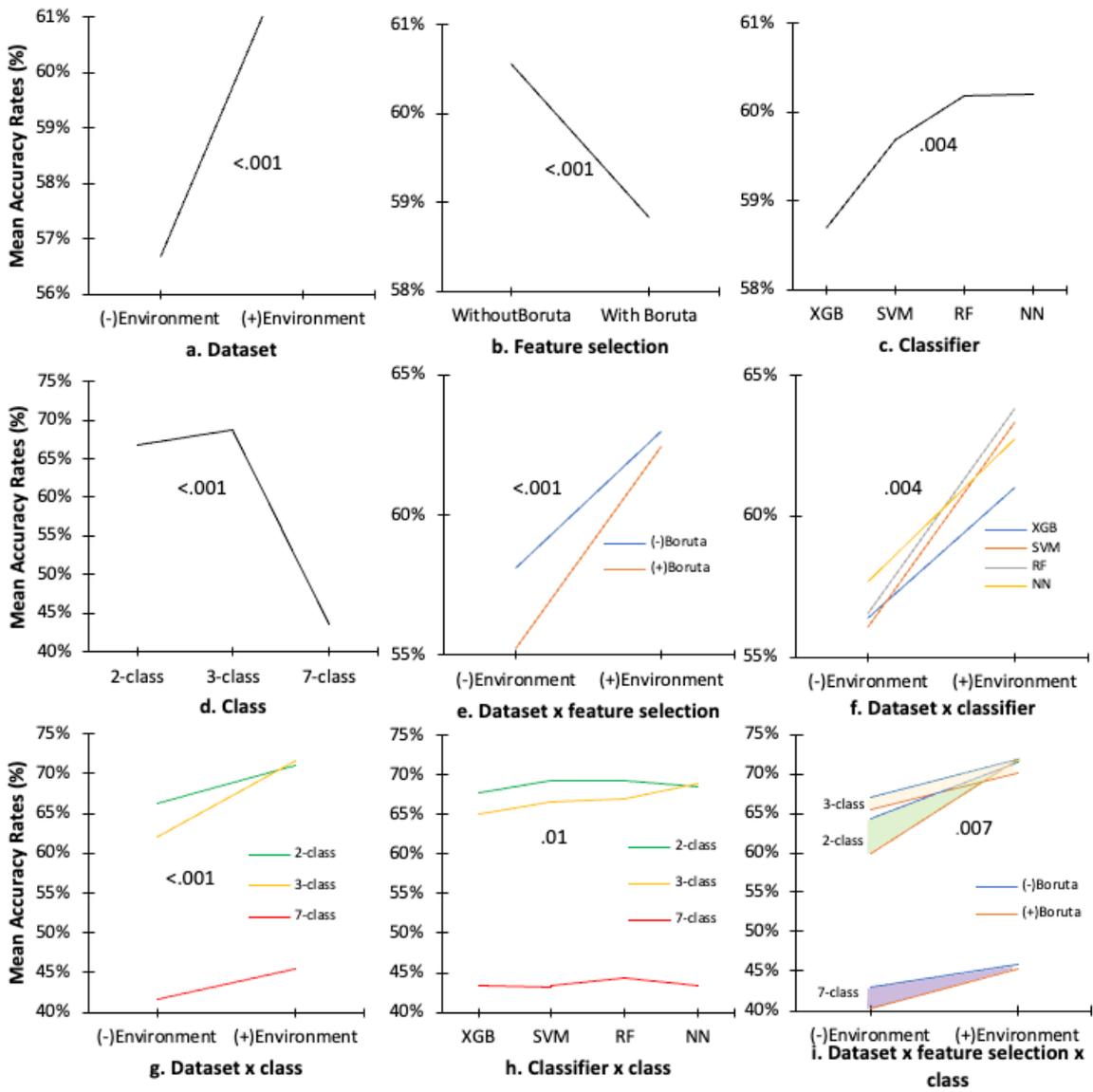

**Figure 7.** Three-way ANOVA results of the differences in the pooled mean classification accuracy rate (%) by dataset, feature selection, classifier, and class

## 4. Discussion

In this study, we were able to categorize and discriminate the behavior of children with PIMD/SMID to binary and multi-class (3 and 7) behavior outcomes. Creating and combining different datasets to investigate whether the inclusion of environment data and conducting feature selection method would allow more accurate classification performance of four different machine learning predictive models in classifying major and minor categories of the behaviors to binary and multiclass behavior outcomes was also achieved. We were also able to evaluate the influences of and the interaction among the behavior categories, dataset, feature selection, and different machine learning classifiers in increasing classification accuracy rates. Our study introduces an interesting, significant, and novel perspective on investigating the importance of environment data especially weather parameters in classifying the behavior of children with PIMD/SMID. Our study also extends previous investigations on the use of machine learning models and the Boruta feature selection method in analyzing and classifying the behavior of children with PIMD/SMID by utilizing behavior data directly observed and collected from the interaction and communication of child-caregiver dyads.

The combination of child characteristics with major or minor (and both) behavior categories obtained the significantly highest classification accuracy rates in classifying binary and 3 behavior outcome classes. Dataset, feature selection, classifier, and class had significant effects on the overall mean classification accuracy rates with high variances of 98% and 79% were attributed to dataset and class, respectively. This indicates higher classification accuracy rates in the dataset with environment data, non-Boruta trained, SVM, RF and NN classifiers, and class 3 ($P<.001$), respectively. These classifiers also had the highest recall/sensitivity of 56% (SVM and RF). However, RF and NN had the highest specificity (79% and 80%, respectively), precision (62%), AUC (71%), and F1 score of 61% (RF) among the classifiers. The interactions between dataset with feature selection, between dataset and classifiers, and between dataset and classes were found to have significant effects on the mean classification accuracy rates ($P<.001$, $P = 0.004$, and $P<.001$, respectively). Non-Boruta trained dataset with environment data, a dataset with environment data using RF classifier, and a dataset with environment data in classifying binary behavior outcomes had relatively higher mean classification accuracy rates. Moreover, the interaction between classifier and class ($P = 0.01$) and the interactions among dataset, feature selection, and class ($P = 0.007$) also had significant effects on the mean classification accuracy rates which revealed that the use of NN classifier in binary classification, SVM, and RF in class 3 and Boruta-trained dataset with environment data in classifying binary behavior outcomes and non-Boruta dataset with environment data in class 3 also had significantly higher mean classification accuracy rates.

Varying classification accuracy rates were found in different recalibrated dataset combinations. Recalibration of the dataset has been introduced previously and findings revealed that classification accuracy could be increased when using data from the beginning of the session were used to recalibrate the classifier [27]. Although recalibration of the dataset was not performed by dividing each session into data blocks for cross-validation of the classification and tested different decoding schemes, in our study, combining child characteristics with either major or minor behavior categories or both with environment data have improved the classification accuracy rates [27]. High accuracy rates in classifying binary and 3 behavior class outcomes regardless of behavior category (minor, major, or both) support the results that 98% of the variances to the mean classification accuracy rates were attributed to the inclusion of environment data in the dataset combinations. The inclusion of environment data and its interaction with feature selection to the binary and 7-class and overall mean classification accuracy rates also revealed that dataset combination with environment data had higher classification rates either trained with Boruta or not than dataset without environment data.

Our hypothesis that training the datasets with feature selection using the Boruta algorithm would improve classification accuracy performance partially supports our findings on non-Boruta trained datasets having significantly higher mean classification accuracy rates than Boruta-trained datasets as revealed by its significant effect on the overall mean classification accuracy rates. As previously mentioned, a similar study that investigated training data with Boruta feature selection to improve classification performance found that the classifier with feature selection was higher (82%) than that of the dataset trained without feature selection (74%) [28]. Although a parallel result of higher specificity and sensitivity were observed in the model with feature selection, the two classifiers were less different based on the overall accuracy [28]. Our study also revealed contrasting results in terms of the significant interaction between feature selection, dataset, and class. While Boruta-trained dataset with environment data in classifying binary behavior outcomes had a higher mean classification accuracy rate, a non-Boruta dataset with environment data in class 3 also had significantly higher mean classification accuracy rates. These could provide evidence that the performance of classifiers trained with feature selection using Boruta was not affected by the dataset combination as both had environment data. However, it might be sensitive to the number of classification outcomes and that other feature selection methods could best fit in training. Although our study did not compare Boruta with other feature selection methods, the results of a study by Chen (2020) could explain in part the higher accuracy rates of non-Boruta trained datasets than that of Boruta-trained ones in classifying multiclass outcomes [39]. When the results of the dataset with and without important features selection by RF methods varImp(), Boruta, and RFE were compared to get the best accuracy, Boruta was not the best feature selection method [39]. The performance evaluation in all their experiments with three different dataset methods identified varImp()RF as the best classifier [39].

Overall, the mean classification accuracy rate of datasets analyzed using SVM, RF and NN classifiers had significantly higher mean classification accuracy rates than the dataset analyzed using the XGB classifier. While RF and SVM had the highest recall/sensitivity, RF and NN had the highest specificity, precision, AUC, and F1 score (RF) among the classifiers. This supports our hypothesis partially in terms of SVM-based classification with its high recall/sensitivity rates however, RF had the better performance as it not only had the highest high recall/sensitivity rates but also obtained the highest specificity, precision, AUC, and F1 score. Several studies that compared the performance of RF with other classifiers that were also used in this study such as NN and SVM found similar results. A regression study aimed to compare supervised machine learning methods like RF, SVM, NN, and LM in terms of the lower margin of error. Using mean absolute error and root mean square based on test-set, RF has the lowest error as revealed by the highest R-square value of 96% compared with other models which obtained R-square of 93% [40]. This could be because RF, compared with other models that perform better in bigger datasets, can identify patterns and consequently producing minimal errors especially in handling small datasets like ours. Similarly, the RF method was also found to be the best classifier method than accuracy than other classifiers including SVM [40]. RF as a decision tree classification tool is efficient in handling both nominal and continuous attributes and insensitivity to data distribution assumptions with 93.36%, 93.3%, and 98.57% accuracy rates with 4, 6, and 561 features, respectively [40]. On the other hand, it can be noticed that RF's performance increases as the number of features increases as it relies on multiple decision trees, unlike SVM which builds a model on hyperplanes based on support vectors [40]. SVM had a high accuracy rate of 96% with only 5 features than RF that had an accuracy of 95% but utilizing 9 features [21]. RF was also compared with NN, although accuracy rates were not significantly different, NN, on average, obtained higher accuracy rates than RF [23].

Interestingly, our results also showed that XGBoost may not be suitable for the classification task at hand despite its proven performance as a machine learning problem solver [41]. A study that investigated its parameter tuning process, and compared its performance by training speed and accuracy with gradient boosting and RF found that XGB and gradient boosting performed the least compared with RF [41]. The difference in the performances of XGB and RF was attributable to the use of the default parameter [41]. While XGB and gradient boosting did not perform well using the default parameter thus require parameter search to create accurate models based on gradient boosting, RF, on the other hand, performs better when the default parameter values were used [41]. Further, the study also concluded that in randomization, setting the subsampling rate and changing the number of features to sqrt selected at each split to reduce the size of the parameter grid to 16-fold and improving the classifier's average performance was not necessary [41].

Significantly higher overall classification accuracy rates were obtained in classifying 3-class behavior outcomes compared with classifying binary (2-class) and 7-class behavior outcomes. In a similar previous study that explored the classification of several functional reaching movements from the same limb using EEG oscillations to create a versatile BCI for rehabilitation, decreasing accuracy rates (67%, 62.7%, and 50.3%) were observed in decoding 3, 4 and 6 tasks outcomes from the same limb, respectively [27]. However, the difference between binary and multiclass classification should also be investigated with its interaction with the classifiers and as mentioned, classifiers perform differently in terms of the number of classification outcomes and data partition [42]. Jha et al. (2019) also explored and compared the performances of several classifications including but not limited to SVM and RF in improving the result of binary class and multi-class problems [42]. In binary classification, the RF algorithm performs better than other models with a 99.29% accuracy rate, highest precision and highest F-Score applied to (80, 20) % partition on a train-test dataset which was similar to our study [42]. However, when it comes to multiclass classification, Decision Tree with CTREE value consistently had high accuracy (91.11% and 90%) in both (70, 30) % and (80, 20) % dataset partition [42]. Since we did not test the classifiers' performances in terms of dataset partition, classifiers RF and NN, compared with other models, perform differently in terms of the number of classification outcomes which is supported by our findings that the interaction between the classifiers and class was also found to have a significant effect to the variance in the mean classification rates where better performances were found in datasets trained using NN classifier in binary class, and SVM and RF classifiers in 3-class. The differences in the classification accuracies were also dependent on its interaction with dataset combinations where we found that datasets with environment data in the binary class had relatively higher mean classification accuracy rates than datasets with environment data in multiclass behavior outcomes.

The feasibility of the models presented in this study should be further investigated and experiments must be performed to address several limitations. First, the results are limited on the behavior of children with PIMD/SMID in a school setting attending one single special school. Investigating these children in other settings and including those who are attending regular schools or healthcare facilities should be considered by future investigations. Further, although this study included a relatively higher number of children with PIMD/SMID or severe or profound IDs (n = 20), we only examined a collective categorization of the behavior and their classification to different outcomes which is also an important limitation thus needs to be addressed. Due to the very distinctive and unique behaviors of children with PIMD/SMID, as such, an investigation would require analysis at an individual level. Future investigation should consider investigating the classification performances per subject rather than by group to see whether there would be a significant improvement for the system to be optimized per individual. Another limitation is related to testing the hypothesis that feature selection would increase the accuracy rates of the classifiers and the reliance on training the dataset using the Boruta algorithm only. As mentioned in the discussion, future studies should compare the accuracy rates using other feature selection methods like RF methods varImp() and RFE. Lastly, this study was limited to the use of RF, SVM, NN, and XGB in classifying binary and multiclass behavior classification outcomes. The performances of other classifiers should also be compared to introduce different perspectives on the feasibility of using other classifiers in classifying the behavior of children with PIMD/SMID.

## 5. Conclusion

Our study demonstrated the feasibility of classifying the behavior of children with PIMD into binary and multiclass outcomes and relatively higher accuracy rates can be obtained regardless of dataset combination of child characteristics with major or minor (and both) behavior categories, and the inclusion of environment data. Moreover, the improvement in the overall classification accuracy rates is also dependent on the interaction between the classifier and classes and the interactions among dataset, feature selection method, and classes. The use of NN classifier in binary classification, SVM, and RF in class 3 and Boruta-trained dataset with environment data in classifying binary behavior outcomes and non-Boruta dataset with environment data in class 3 could significantly increase the overall mean classification accuracy rates. This highlights that although our results may be promising, classification accuracy performance could still be significantly improved to achieve optimal performance for the system that we are developing. Most importantly, this study could provide the evidence that classifying the behavior of children with neurological and/or motor and physical impairments can be potentially conducted not just for disorder-related evaluation and/or pre-clinical screening and triage, diagnosis, and treatment and intervention but also for support and intervention to achieve independent mobility and communication.

**Supplementary Materials:**
**Supplementary File S1.** Classification accuracy rates of all the algorithms in each dataset combination

**Acknowledgments**

This study was supported by the Ministry of Internal Affairs and Communications' Strategic Information and Communications R&D Promotion Programme (SCOPE) (MIC contract number: 181609003; research contract number 7; 2019507002; 2020507001) which had no role in the design, data collection and analysis, and writing of this manuscript. The authors would like to thank all the children and their caregivers who participated in this study.

**Author Contributions**

**Von Ralph Dane Marquez Herbuela**: Validation, Formal analysis, Writing-Original Draft and Review and Editing, Visualization **Tomonori Karita**: Conceptualization, Methodology, Funding acquisition, Project administration, Supervision, Writing-Review and Editing, Resources **Yoshiya Furukawa**: Methodology, Validation, Formal Analysis, Investigation, Data Curation, Writing-Review and Editing **Yoshinori Wada**: Validation, Formal analysis, Data Curation, Writing-Review and Editing **Shuichiro Senba, Eiko Onishi, Tatsuo Saeki:** Conceptualization, Software, Resources, Project administration, Writing-Review and Editing

**Conflict of Interest**
None declared.